\documentclass{article}
\PassOptionsToPackage{numbers,sort&compress}{natbib}
\usepackage[preprint]{neurips_2026}
\usepackage[utf8]{inputenc}
\usepackage[T1]{fontenc}
\usepackage{amsmath,amssymb,amsthm}
\usepackage{mathtools}
\usepackage{bm}
\usepackage{booktabs}
\usepackage{array}
\usepackage{multirow}
\usepackage{graphicx}
\usepackage{algorithm}
\usepackage{algpseudocode}
\usepackage{float}
\usepackage{hyperref}
\hypersetup{hidelinks}
\usepackage{xcolor}
\usepackage{siunitx}
\usepackage{url}
\usepackage[protrusion=true,expansion=false]{microtype}
\usepackage{fvextra}
\usepackage{tikz}
\usetikzlibrary{arrows.meta,positioning,fit,backgrounds}
\usepackage{subcaption}
\usepackage{tabularx}
\newcolumntype{L}[1]{>{\raggedright\arraybackslash}p{#1}}

\newcommand{\R}{\mathbb{R}}

\newcommand{\E}{\mathbb{E}}

\newcommand{\topk}{\mathrm{TopK}}
\newcommand{\sign}{\mathrm{sign}}

\newcommand{\node}{\mathcal{V}}
\newcommand{\edge}{\mathcal{E}}
\newcommand{\graphs}{\{G_s\}_{s=1}^{S}}
\newcommand{\effecti}[2]{E_{#1}^{(#2)}}
\newcommand{\patch}[2]{\tilde{f}_{#1}(#2)}

\title{Patch-Effect Graph Kernels for LLM Interpretability}

\author{%
  \begin{minipage}[t]{0.45\linewidth}
    \centering
    Ruben Fernandez-Boullon \\
    \texttt{ruben.fernandez.boullon@uvigo.gal}
  \end{minipage}%
  \hfill
  \begin{minipage}[t]{0.45\linewidth}
    \centering
    David N. Olivieri \\
    \texttt{olivieri@uvigo.gal}
  \end{minipage} \\[1.2em]
  Department of Computer Science, University of Vigo, Ourense 32004, Spain
}

\begin{document}
\raggedbottom
\maketitle

\begin{abstract}
Mechanistic interpretability aims to reverse-engineer transformer computations by identifying causal circuits through activation patching. However, scaling these interventions across diverse prompts and task families produces high-dimensional, unstructured datasets that are difficult to compare systematically. We propose a framework that reframes mechanistic analysis as a graph machine-learning problem by representing activation-patching profiles as patch-effect graphs over model components. We introduce three graph-construction methods—direct-influence via causal mediation, partial-correlation, and co-influence—and apply graph kernels to analyze the resulting structures. Evaluating this approach on GPT-2 Small using Indirect Object Identification (IOI) and related tasks, we find that patch-effect graphs preserve discriminative structural signals. Specifically, localized edge-slot features provide higher classification accuracy than global graph-shape descriptors. A screened paired-patching validation suggests that CI- and PC-selected candidate edges correspond to stronger activation-influence effects than random or low-rank candidates. Crucially, by evaluating these representations against rigorous prompt-only and raw patch-effect controls, we make the evidential scope of the benchmark explicit: graph features compress structured patching signal, while raw tensors and surface cues define strong baselines that any circuit-level claim should address. Ultimately, our framework provides a compression and evaluation pipeline for comparing patching-derived structures under controlled baselines, separating robust slice-discriminative evidence from stronger task-general causal-circuit claims.
\end{abstract}

\section{Introduction}
\label{sec:intro}

Transformer models \cite{vaswani2017attention,radford2019language} distribute computation across layers, tokens, attention heads, and MLP sublayers. Mechanistic interpretability aims to reverse-engineer this computation by identifying the circuits---structured subgraphs of internal components---responsible for specific behaviours \cite{elhage2021transformer}. The standard tool for causal discovery within such circuits is \emph{activation patching} \cite{wang2023localizing}: given a clean prompt and a corrupted prompt designed to break a target behaviour, one replaces the activation of a chosen internal node during the corrupted forward pass with the corresponding clean value and measures the change in a scalar observable (e.g.\ the logit of the target token). A large positive patch effect indicates that the node causally restores the target behaviour.

A single patching experiment yields one scalar effect size. Repeating it across all nodes, all examples in a prompt family, and multiple corruption types produces a high-dimensional \emph{interventional dataset} that reflects how the chosen observable responds to interventions. Three fundamental difficulties prevent direct use of this raw tensor:
\begin{itemize}
  \item How do we compress patch-effect profiles into structured, comparable objects that respect the relational geometry of the network?
  \item How do we measure similarity between circuits discovered under different tasks or corruptions?
  \item Can a principled similarity geometry reveal when circuits are shared, distinct, or robust to noise?
\end{itemize}

This paper proposes a pipeline that addresses all three difficulties simultaneously. The key insight is that each \emph{slice}---a (task family, corruption type) pair---naturally defines one graph over a fixed node set $\node$ (layers $\times$ tokens $\times$ component types), where edge weights encode relationships between patch-effect profiles across examples. A collection of $S$ such graphs $\graphs$ is a \emph{graph dataset} on which we can perform supervised learning. This reframes mechanistic analysis as a \emph{graph machine-learning} problem: learn a similarity geometry across patching-derived graphs, not just within a single prompt.

\begin{figure}[H]
\centering
\begin{tikzpicture}[
  box/.style={draw, rounded corners=3pt, fill=blue!8, minimum height=1.0cm,
              text width=2.35cm, align=center, font=\small},
  arr/.style={-{Stealth[length=5pt]}, thick},
  note/.style={font=\scriptsize\itshape, text=gray, align=center, text width=2.35cm}
]
\node[box] (tf)    {Frozen\\transformer $f_0$};
\node[box, right=0.8cm of tf]  (pat)  {Activation\\patching};
\node[box, right=0.8cm of pat] (ten)  {Effect tensors $E_u^{(i)}$};
\node[box, right=0.8cm of ten] (gr)   {Graph dataset $\{G_s\}$};
\node[box, below=1.25cm of gr] (emb)  {Graph\\embedding $x_s$};
\node[box, left=0.8cm of emb]  (ker)  {Kernel\\$K_{st}$};
\node[box, left=0.8cm of ker]  (svm)  {SVM / clustering};

\draw[arr] (tf)  -- (pat);
\draw[arr] (pat) -- (ten);
\draw[arr] (ten) -- (gr);
\draw[arr] (gr)  -- (emb);
\draw[arr] (emb) -- (ker);
\draw[arr] (ker) -- (svm);

\node[note, below=0.28cm of pat] {clean $\to$ corrupted};
\node[note, below=0.28cm of ten] {per node, per example};
\node[note, above=0.22cm of gr]  {one per slice; DI, PC, or CI};
\node[note, below=0.28cm of emb] {WL, spectral, graphlet};
\node[note, below=0.28cm of ker] {linear or RBF};
\end{tikzpicture}
\caption{End-to-end pipeline from activation patching to kernel-based circuit classification. A frozen transformer and controlled prompt families feed a patching engine; effect tensors are aggregated into one graph per slice; graphs are embedded and compared via a kernel; downstream classification and clustering happen across graphs, not within a single prompt.}
\label{fig:pipeline}
\end{figure}
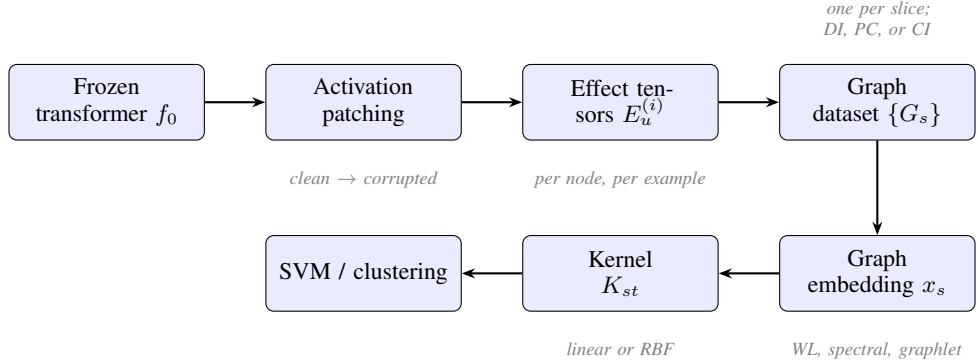

Figure~\ref{fig:pipeline} illustrates the end-to-end pipeline. The same frozen transformer and prompt families feed a patching engine that outputs effect tensors $\effecti{u}{i}$ per node per example. These tensors are aggregated into graphs (one per slice) via one of three construction methods. Each graph is embedded by a graph feature map $\phi_\mathrm{graph}$, producing a vector $x_s \in \R^d$. A kernel $K_{st} = k(x_s, x_t)$ then defines a similarity geometry on which an SVM performs downstream classification.

We make three principal contributions.
\begin{enumerate}
  \item We introduce \emph{interpretability graphs}: graph representations built from activation-patching profiles, where each slice of the interventional dataset maps to one graph with a fixed node set and slice-specific weighted edges. Learning happens \emph{across bootstrap graphs}, not within a single prompt.
  \item We define three edge-construction methods—direct influence, partial correlation, and co-influence—and empirically compare the two scalable approximations, CI and PC, under a common IOI protocol. We additionally validate DI through a screened paired-patching study on CI/PC-selected edges, leaving exhaustive DI scaling as the natural next stage.
  \item By comparing graph representations against raw patch-effect, prompt-only surface-cue, and learned graph-encoder (GCN+SVM) baselines, we stress-test current mechanistic evaluation standards. We show how prompt-only controls reveal surface shortcuts in the historical IOI comparison, how raw patch-effect tensors provide a strong calibration point for graph-compression claims, and how localised edge-slot identity transfers across model sizes (GPT-2 small and DistilGPT-2) while subtree-based compressions are model-size dependent.
\end{enumerate}

\section{Related Work}
\label{sec:related}

The mechanistic interpretability programme has identified circuit-level mechanisms in transformer models, including induction heads \cite{olsson2022context}, indirect object identification \cite{wang2023localizing}, and greater-than circuits \cite{hanna2023does}. Our work extends this programme by providing a systematic, kernel-based framework for \emph{comparing} circuits across tasks and corruptions rather than characterising each circuit independently.

Activation patching \cite{wang2023localizing} and its variants---path patching \cite{goldowsky2023localizing}, attribution patching \cite{nanda2023attribution}, and causal mediation analysis \cite{pearl2009causality}---have become standard tools for causal analysis in interpretability. A single-node patch yields a scalar effect $E_u$; \emph{paired} sequential patching of two nodes $u$ then $v$ yields a joint effect $E_{u,v}$ whose excess over $E_u$ measures the causal mediation of $u$'s restoration by $v$. Our direct-influence graph is designed to use this mediation signal as the edge weight, making it the causal target construction; we empirically study scalable CI/PC constructions and validate DI through screened paired patching.

Graph kernels provide a principled way to define similarity between graphs without explicit graph isomorphism testing \cite{kriege2020survey}. The Weisfeiler--Lehman (WL) subtree kernel \cite{shervashidze2011weisfeiler} is a strong baseline that compresses each graph into a histogram of local subtree patterns. Spectral kernels use eigenvalue decompositions of the graph Laplacian \cite{kondor2002diffusion}. Graphlet kernels count occurrences of small subgraph motifs \cite{shervashidze2009efficient}. Our work applies these kernels to interpretability graphs, where node labels carry mechanistic meaning (layer, token, component type).

Scaling interpretability methods to larger models is a persistent challenge \cite{conmy2023towards,bricken2023towards}. Our pipeline addresses scalability through top-$k$ edge sparsification and fixed-dimension graph embeddings (WL, spectral, graphlet) that do not grow quadratically with model size.

\section{Methodology}
\label{sec:method}

\subsection{Formal Setup}

Let $f_0$ be a frozen transformer with $L$ layers, $T$ token positions, $H$ attention heads per layer, and vocabulary $V$. For a prompt $x = (x_1, \dots, x_T)$, write $\ell(x) \in \R^{|V|}$ for the final logit vector. We choose the observable $O(x) = \ell_{y^\star}(x)$, the logit of the target token $y^\star$ at a designated position.

Let $\mathcal{K} = \{\mathrm{res}, \mathrm{att}, \mathrm{mlp}\}$. Define the node set as a disjoint union:
\begin{equation}
  \node = \node_\mathrm{res} \,\dot\cup\, \node_\mathrm{mlp} \,\dot\cup\, \node_\mathrm{att},
\end{equation}
where $\node_\mathrm{res} = \{(\ell, t, \mathrm{res}) : 0 \leq \ell < L,\; 1 \leq t \leq T\}$ and analogously for MLP and attention. The node set $\node$ is \emph{fixed} across all slices; only edge weights vary.

A \emph{slice} $s$ is a (task family, corruption type) pair. For each slice $s$, we sample $N$ paired examples $\{(x_i^\mathrm{cln}, x_i^\mathrm{crp}, y_i^\star)\}_{i=1}^N$, where $x_i^\mathrm{crp}$ is designed to break the target behaviour. The historical IOI comparison is not fully surface-balanced; we therefore make this confound explicit and evaluate it with a prompt-only surface-cue control.

\subsection{Activation Patching}

For each example $i$ and node $u \in \node$:
\begin{equation}
  \effecti{u}{i} = O\bigl(\patch{f_0}{x_i^\mathrm{crp};\, u}\bigr) - O(f_0(x_i^\mathrm{crp})),
\end{equation}
where $\patch{f_0}{x_i^\mathrm{crp};\, u}$ denotes the forward pass on $x_i^\mathrm{crp}$ with node $u$ patched to its clean-run value. A large positive $\effecti{u}{i}$ indicates causal restoration of the target behaviour.

For an ordered pair $(u, v)$ with $u \prec v$:
\begin{equation}
  \effecti{u,v}{i} = O\bigl(\patch{f_0}{x_i^\mathrm{crp};\, u, v}\bigr) - O(f_0(x_i^\mathrm{crp})),
\end{equation}
where both $u$ and $v$ are simultaneously patched. This enables the direct-influence edge construction below.

\subsection{Graph Construction}
\label{sec:graph}

Each slice $s$ yields one directed weighted graph $G_s = (\node, \edge_s, w_s)$ over the fixed node set. We compare three edge-construction methods.

\subsubsection*{(A) Direct-Influence Graph (Primary)}
The direct-influence (DI) edge weight measures causal mediation: does restoring $v$ additionally change the effect of restoring $u$?
\begin{equation}
  w_s^\mathrm{DI}(u \to v) = \E_{i \in \mathcal{Z}_s}\bigl[\effecti{u,v}{i} - \effecti{u}{i}\bigr], \quad u \prec v.
\end{equation}
A large positive weight means $v$ mediates part of $u$'s causal effect on the observable. This requires $O(|\node|^2)$ forward passes per prompt and is the primary target construction. Section~\ref{sec:e2} therefore evaluates DI through a cost-aware screened candidate validation in which CI/PC-ranked edges are re-evaluated with paired patching.

\subsubsection*{(B) Partial-Correlation Graph}
The partial-correlation (PC) edge weight removes shared upstream causes:
\begin{equation}
  w_s^\mathrm{PC}(u \to v) = -\frac{\Sigma^{-1}_{uv}}{\sqrt{\Sigma^{-1}_{uu} \Sigma^{-1}_{vv}}}, \quad u \prec v,
\end{equation}
where $\Sigma = \mathrm{Cov}\bigl(\{\effecti{\cdot}{i}\}_{i \in \mathcal{Z}_s}\bigr)$, estimated as a ridge-regularised covariance matrix before pseudo-inversion. Cost: $O(|\node|^3)$ offline, no additional forward passes. This construction is implemented and evaluated in Section~\ref{sec:e5}.

\subsubsection*{(C) Co-Influence Graph}
The co-influence (CI) edge weight is the Pearson correlation of patch-effect profiles across examples:
\begin{equation}
  w_s^\mathrm{CI}(u \to v) = \mathrm{corr}\bigl(\{\effecti{u}{i}\}_{i \in \mathcal{Z}_s},\, \{\effecti{v}{i}\}_{i \in \mathcal{Z}_s}\bigr).
\end{equation}
This is the cheapest construction ($O(|\node|)$ additional forward passes per prompt). It captures co-variation rather than mediation, making it a scalable correlational graph construction and the primary construction used in the representation study.

\textbf{Why three constructions?} The comparison DI vs.\ PC vs.\ CI tests whether causal mediation is essential. If DI $>$ CI, genuine causal mediation edges are more informative than raw correlation. If DI $\approx$ CI, the cheap co-influence construction is adequate and paired patching can be skipped. Both outcomes are informative.

We enforce $u \prec v$ (by layer, then token index), preventing spurious backward edges and ensuring graphs respect the temporal information flow of the transformer.

For each node $u$, retain only the top-$k$ outgoing edges by absolute weight:
\begin{equation}
  \edge_s = \bigcup_{u \in \node} \bigl\{(u, v) : v \in \topk_k\bigl(\{|w_s(u \to \cdot)|\}\bigr)\bigr\}.
\end{equation}
Negative weights are retained (sign carries mechanistic information). Top-$k$ sparsification fixes graph size across slices and stabilises downstream kernel computations. Default $k=5$.

\subsection{Graph Embedding}

We embed each graph $G_s$ into a vector $x_s = \phi_\mathrm{graph}(G_s) \in \R^d$ using one of five graph feature maps:

The WL algorithm \cite{shervashidze2011weisfeiler} computes a histogram of local subtree patterns up to depth $H_\mathrm{WL}$. Initial labels encode $(\ell, t, \mathrm{type})$. Subsequent labels hash the current label together with the sorted multiset of signed directed neighbour labels. Dimension grows with the number of distinct subtree patterns observed (typically $\sim 10^3$--$10^4$ for GPT-2).

Since all graphs share the same node set, each possible edge slot $(u, v)$ maps to a fixed coordinate. Three variants: weighted ($w(u\to v)$), binary ($\mathbf{1}[(u,v)\in\edge]$), and signed ($\sign(w(u\to v))$). Dimension: $N(N-1) = 14{,}280$ for GPT-2 with residual-stream-only nodes ($N=120$).

Compute the normalised graph Laplacian eigenvalues and eigenvector statistics. Compact fixed-dimension vector ($d=96$) capturing global graph shape.

Count directed 3-node subgraph motifs. Fixed-dimension vector ($d=38$) capturing local graph motifs.

Map each edge slot to a fixed coordinate via hash modulo $d$, accumulating signed weights. Fixed dimension $d=1{,}024$, eliminating the quadratic growth of full fixed-layout.

\subsection{Kernel Methods}

Given embeddings $\{x_s\}_{s=1}^S$, we compute a kernel matrix $K \in \R^{S \times S}$ and train an SVM for downstream classification.

$K_{st}^\mathrm{lin} = x_s^\top x_t$. Works well when the feature space is already well-structured for linear separation; computationally cheap.

$K_{st}^\mathrm{RBF} = \exp(-\gamma \|x_s - x_t\|^2)$, with $\gamma$ set by median heuristic or cross-validation. Implicitly maps features into an infinite-dimensional Hilbert space; can capture non-linear boundaries.

Both kernels are applied on top of the \emph{same} graph embeddings, so differences in downstream accuracy are attributable to the kernel geometry, not to data processing choices.

\section{Experimental Setup}
\label{sec:experiments}

\subsection{Model}

We use GPT-2 small \cite{radford2019language} as our primary model: $L=12$ layers, $H=12$ attention heads, $d_\mathrm{model}=768$, frozen weights. All experiments are forward passes with hook-based interventions at residual-stream positions ($\mathrm{type} = \mathrm{res}$). The resulting node set has $|\node_\mathrm{res}| = L \times T$ nodes, where $T$ is the prompt length. We additionally apply the same CI-graph protocol to DistilGPT-2 \cite{sanh2019distilbert} ($L=6$, same width, frozen weights) as a cross-model replication, reported in Section~\ref{sec:e3}.

\subsection{Pilot Scope: IOI Task Family}
\label{sec:pilot_scope}

Current experiments cover the \textbf{Indirect Object Identification (IOI)} task \cite{wang2023localizing}: the model must predict the indirect object in sentences such as ``Alice gave Bob a book. Bob thanked \_\_\_.'' We evaluate three corruption types:
\begin{itemize}
  \item \texttt{name\_swap}: replace one name with a distractor, breaking the name-binding signal.
  \item \texttt{abba}: swap both names, redirecting the prediction to the other name.
  \item \texttt{second\_subject\_swap}: replace the subject of the second sentence while matching \texttt{abba}'s target/distractor name counts.
\end{itemize}
Each binary comparison yields $S=2$ slices. The IOI circuit is implemented primarily by name-binding and duplicate-token heads at layers 7--10 in GPT-2 \cite{wang2023localizing}, making it a useful controlled testbed for patch-effect measurement and slice-level graph comparison.

The main benchmark uses the historical pair \texttt{name\_swap} vs.\ \texttt{abba}. We also report a surface-balanced pair, \texttt{abba} vs.\ \texttt{second\_subject\_swap}, where both corruptions contain the target name twice and the distractor name once. These binary comparisons are intended as controlled slice-discrimination tests; task-general evaluation requires broader multi-family suites.

To test whether the protocol transfers beyond binary IOI, we additionally run a preliminary $S=4$ evaluation with four slices: \texttt{ioi:abba}, \texttt{ioi:second\_subject\_swap}, \texttt{induction:token\_swap}, and \texttt{induction\_late:token\_swap}. The two induction-template slices are synthetic repeated-token prompts designed to activate different copy positions. This pilot provides an initial multi-slice stress test and motivates scaling to broader task families.

\subsection{Evaluation Protocol}

We generate $N=100$ paired examples per corruption per seed. Each slice produces 32 bootstrap graphs by resampling 75\% of examples with replacement; 32 additional bootstrap graphs form the test set from disjoint examples. We use seeds $\{7, 42, 123\}$ and report mean $\pm$ population standard deviation across seeds. Classification uses a linear or RBF SVM on training bootstrap graphs; we report accuracy on the example-disjoint test set. Because the two classes are bootstrap resamples of only two underlying IOI slices, these accuracies measure slice-discriminative signal within the current benchmark, not generalisation across independent circuit families.

For the preliminary $S=4$ evaluation, we run a staged grid $N \in \{20,50,100\}$ with the same example-disjoint split but only 8 bootstrap graphs per slice to keep the experiment cheap. The chance baseline is therefore $0.25$. These results are reported separately from the main IOI tables.

We apply edge shuffle (randomly reassign edge targets, preserving the weight multiset) and weight shuffle (permute weights across edges while preserving positions) to the fixed-layout weighted baseline. These controls test whether the signal depends on edge-slot assignments rather than marginal graph statistics.

We also evaluate a raw patch-effect classifier and a prompt-only surface-cue classifier. The raw baseline calibrates how much discriminative signal is already present before graph compression. The surface-cue baseline tests whether \texttt{name\_swap} and \texttt{abba} are separable from prompt text alone.

\section{Results}
\label{sec:results}

\subsection{E1: Patch-Effect Tensor Validation}
\label{sec:e1}

Confirm that patch-effect tensors $\{\effecti{u}{i}\}_{u,i}$ carry structured, non-random signal before graph construction.

We compute mean patch effects $\bar{E}_u = \E_i[\effecti{u}{i}]$ for each node $u$ over $N=8$ representative examples per corruption and visualise them as heatmaps over the layer--token grid. The full $N=100$ examples per corruption are used in the classification experiments of Section~\ref{sec:e3}.

\begin{table}[!ht]
\centering
\small
\begin{tabular}{lccc}
\toprule
\textbf{Slice} & \textbf{Grid} ($L \times T$) & \textbf{Scale $|\bar{E}_u|_{99}$} (logit) & \textbf{Examples} \\
\midrule
IOI: name\_swap & $12 \times 11$ & $3.23$ & $8$ \\
IOI: abba       & $12 \times 11$ & $5.48$ & $8$ \\
IOI: second\_subject\_swap & $12 \times 10$ & $1.25$ & $8$ \\
\bottomrule
\end{tabular}
\caption{Patch-effect heatmap statistics for the IOI pilot (GPT-2 small, residual-stream nodes, $N=8$ examples per slice used for visualisation). The scale column is the robust symmetric colour limit used by each heatmap. Signal concentrates in layers 5--10, consistent with the known IOI circuit \cite{wang2023localizing}.}
\label{tab:e1}
\end{table}

\begin{figure}[!ht]
\centering
\captionsetup[subfigure]{font=scriptsize,skip=1pt}
\begin{subfigure}[b]{0.32\textwidth}
  \includegraphics[width=\textwidth]{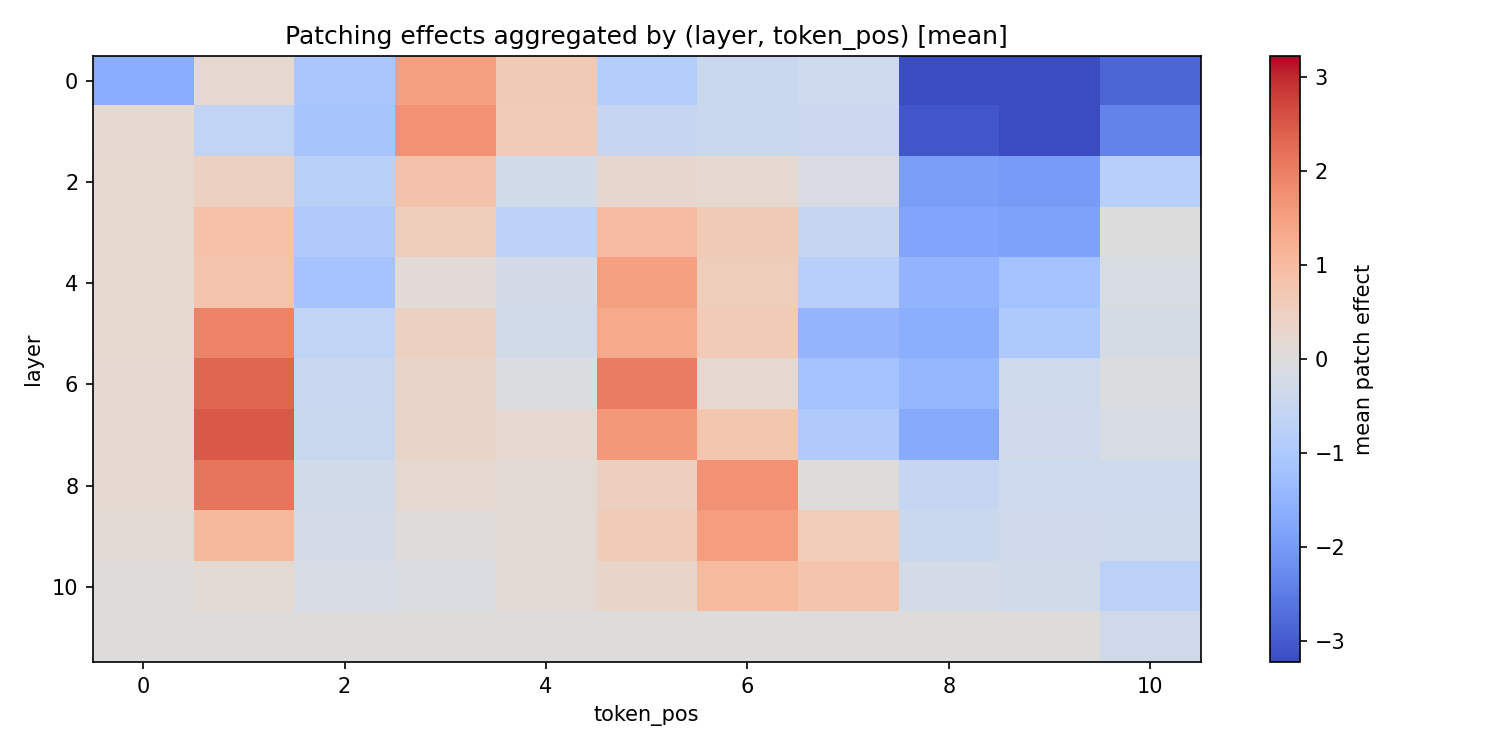}
  \caption{\texttt{name\_swap}}
\end{subfigure}
\hfill
\begin{subfigure}[b]{0.32\textwidth}
  \includegraphics[width=\textwidth]{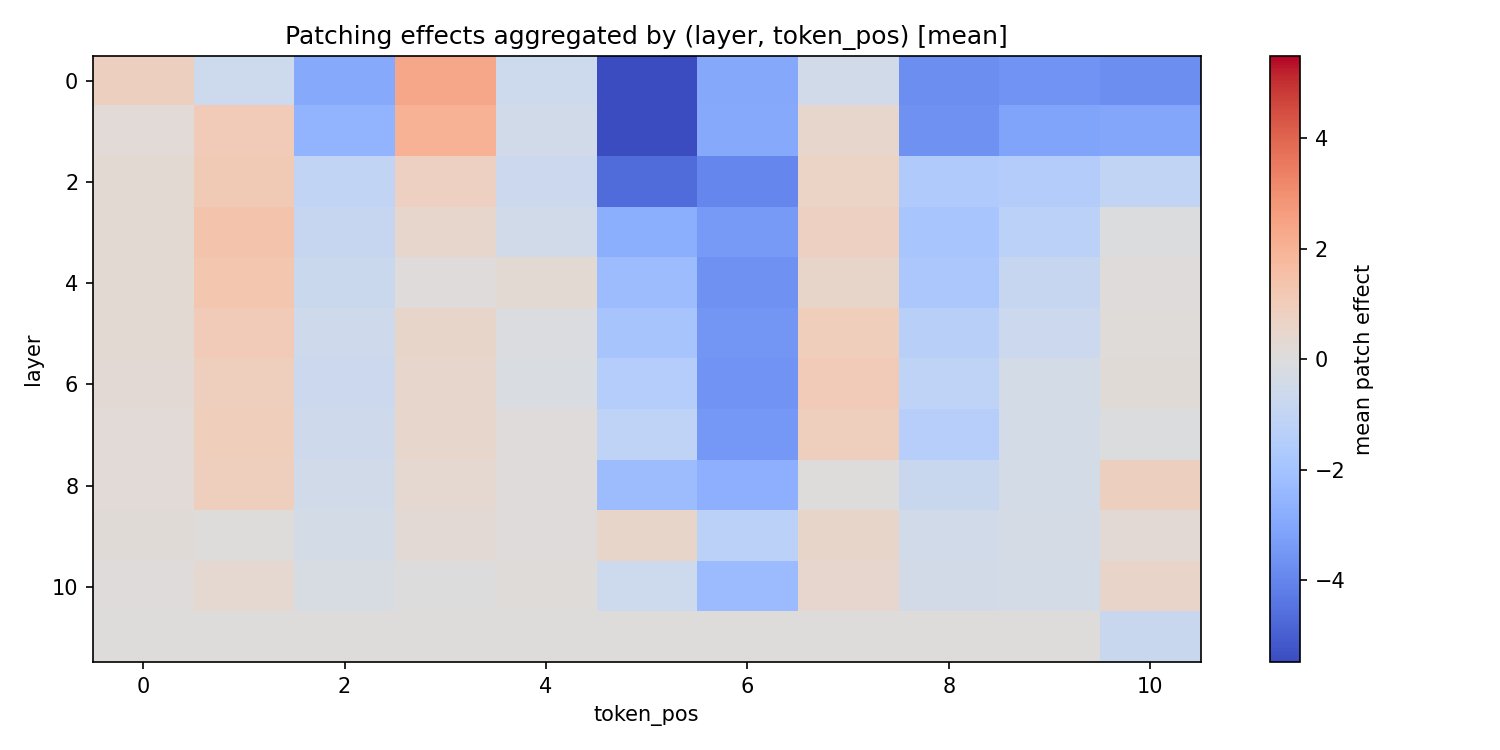}
  \caption{\texttt{abba}}
\end{subfigure}
\hfill
\begin{subfigure}[b]{0.32\textwidth}
  \includegraphics[width=\textwidth]{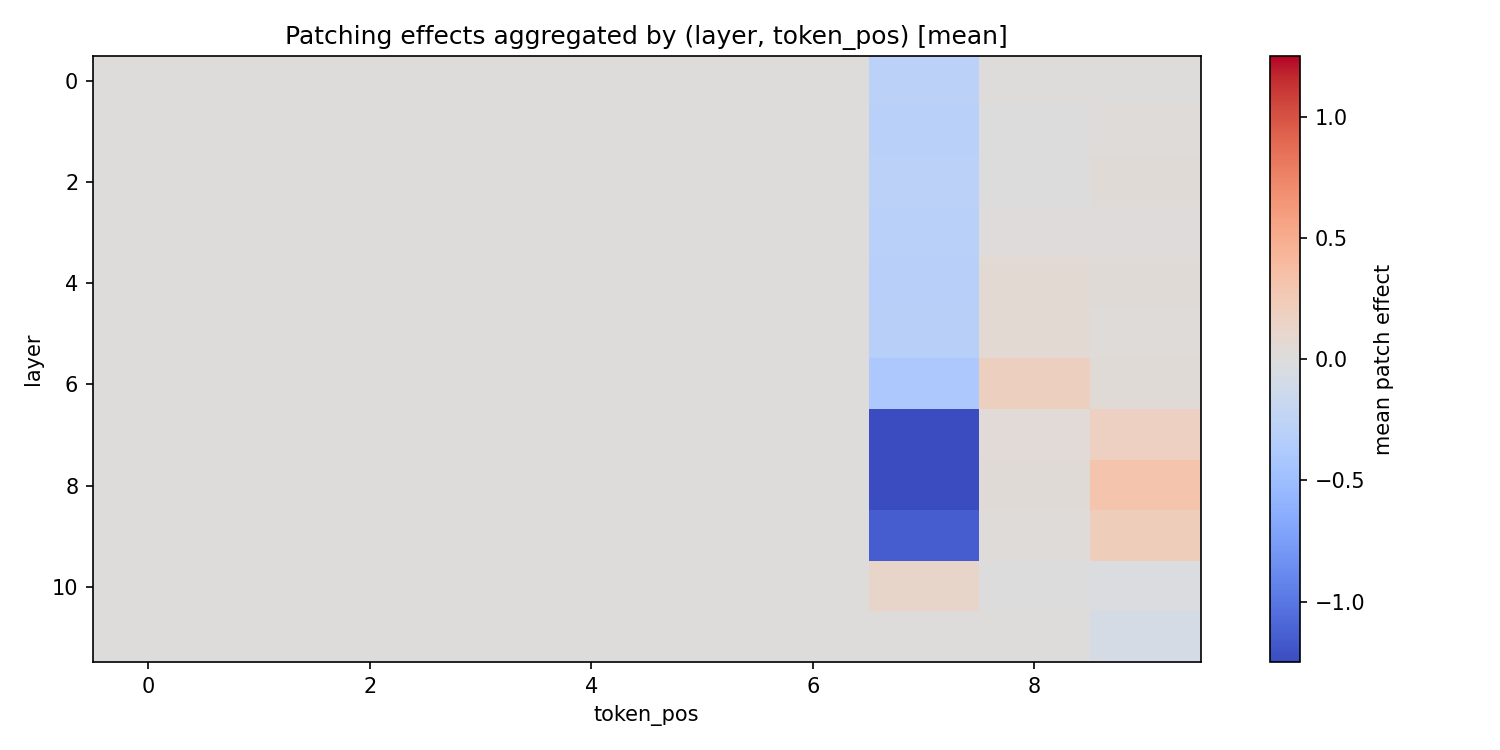}
  \caption{\texttt{2nd-subj}}
\end{subfigure}
\caption{Mean patch-effect heatmaps for GPT-2 small residual-stream nodes ($N=8$ selected examples per slice). Colour encodes the mean patch effect in logit units (red: positive; blue: negative). The added \texttt{second\_subject\_swap} panel shows the surface-balanced IOI control used later in the classification benchmarks.}
\label{fig:heatmaps}
\end{figure}

Figure~\ref{fig:heatmaps} shows the mean patch-effect heatmaps for the historical IOI corruptions and the surface-balanced control. The slices show structured signal with different spatial signatures: \texttt{name\_swap} concentrates positive effects at early token positions and at the last two tokens; \texttt{abba} shifts high-magnitude signal toward middle token positions and shows a distinct negative band; \texttt{second\_subject\_swap} is lower-amplitude but still structured. These heatmaps establish structured slice signal; later controls separate this signal from prompt-surface effects.

Heatmaps show structured bands concentrated in specific layers and token positions, consistent with the known IOI circuit. Random nodes produce near-zero effects on average.

We run a small paired-node causal evaluation on GPT-2 (seed $42$, $N=10$ paired prompts, 10 candidate edges). All 10 candidate edges survive FDR correction (BH method, $\alpha=0.05$), with top edges concentrated in the last-token residual stream at layers 9--11 (e.g.\ $\ell=9, t=T_\mathrm{last}$ and $\ell=11, t=T_\mathrm{last}$), consistent with known IOI save-slot positions. This provides targeted edge-level causal validation for the screened candidates.

\subsection{E2: Graph Construction Comparison}
\label{sec:e2}

CI and PC graph construction are implemented and evaluated under the same GPT-2 IOI protocol in Section~\ref{sec:e5}. DI is the computationally expensive causal target construction, requiring paired-node patching at $O(|\node|^2)$ passes per prompt. We therefore evaluate it through screened candidate validation: CI/PC propose candidate edges, and paired patching re-evaluates those edges on held-out prompts.

Appendix Table~\ref{tab:screened_di} reports a GPT-2 screened-DI pilot on the surface-balanced IOI pair (\texttt{abba} vs.\ \texttt{second\_subject\_swap}). We generate $N=100$ examples per slice, use a stratified discovery/evaluation split with 50 examples in each pool, rank residual-stream candidate edges on the discovery pool, and evaluate 50 held-out edges per group with paired patching. The level-A activation-influence score $I(u\to v)$ is larger for CI- and PC-screened candidates ($0.373$ and $0.305$) than for random or low-rank candidates ($0.122$ and $0.102$). Mediation $M$ and necessity are more variable, so the pilot supports screened causal activity while reserving exhaustive DI graph benchmarking for the scaling study.

The current graph-construction comparison therefore isolates two questions: whether removing linear shared-cause structure via PC improves classifiability over cheap CI, and whether CI/PC-ranked candidates survive paired-patching validation better than simple null screens. Exhaustive DI-vs-correlational graph benchmarking is the direct follow-up once paired-patching cost is scaled.

\subsection{E3: Representation Baselines (CI Graphs, IOI Binary)}
\label{sec:e3}

Establish representation baselines on co-influence (CI) graphs and compare them against non-graph controls.

GPT-2 small, IOI task, $S=2$ slices (\texttt{name\_swap} vs.\ \texttt{abba}), $N=100$ examples per slice, $k=5$, residual-stream nodes ($|\node|=120$), seeds $\{7, 42, 123\}$.

\begin{table}[!ht]
\centering
\small
\begin{tabular}{llccc}
\toprule
\textbf{Representation} & \textbf{Kernel} & \textbf{Acc.\ (mean)} & \textbf{Std} & \textbf{Dim.} \\
\midrule
\multicolumn{5}{l}{\textit{Non-graph controls}} \\
Prompt surface cues & Linear & $1.0000$ & $0.0000$ & $6$ \\
Raw patch-effect tensor & Linear & $1.0000$ & $0.0000$ & $120$ \\
Top-$k$ node identity & Linear & $0.9917$ & $0.0118$ & sparse \\
\midrule
\multicolumn{5}{l}{\textit{Global shape descriptors}} \\
Spectral          & Linear  & $0.5000$ & $0.0000$ & $96$ \\
Graphlet          & Linear  & $0.7604$ & $0.1239$ & $38$ \\
\midrule
\multicolumn{5}{l}{\textit{Subtree pattern features}} \\
WL subtree        & Linear  & $0.9115$ & $0.0321$ & ${\sim}18{,}500$ \\
WL subtree        & RBF     & $0.6146$ & $0.1620$ & ${\sim}18{,}500$ \\
\midrule
\multicolumn{5}{l}{\textit{Localised edge-slot features}} \\
Fixed layout weighted & Linear & $0.9688$ & $0.0338$ & $14{,}280$ \\
Fixed layout binary   & Linear & $0.9896$ & $0.0147$ & $14{,}280$ \\
\textbf{Fixed layout signed}   & Linear & $\mathbf{1.0000}$ & $\mathbf{0.0000}$ & $14{,}280$ \\
Hashed sign           & Linear & $0.9948$ & $0.0074$ & $1{,}024$ \\
Hashed weighted       & Linear & $0.9635$ & $0.0295$ & $1{,}024$ \\
Coarse count          & Linear & $0.7552$ & $0.1278$ & ${\sim}107$ \\
\midrule
\multicolumn{5}{l}{\textit{Learned graph encoder}} \\
GNN encoder + SVM     & Linear & $0.6094$ & $0.1230$ & $32$ \\
GNN encoder + SVM     & RBF    & $0.5833$ & $0.0868$ & $32$ \\
\midrule
\multicolumn{5}{l}{\textit{Null controls (fixed layout weighted)}} \\
Edge shuffle      & ---   & $0.5052$ & $0.0074$ & --- \\
Weight shuffle    & ---   & $0.5191$ & $0.0346$ & --- \\
\midrule
Chance baseline   & ---   & $0.5000$ & --- & --- \\
\bottomrule
\end{tabular}
\caption{Representation results on IOI with GPT-2 small ($N=100$, $k=5$, CI graphs, seeds $\{7,42,123\}$, binary classification: \texttt{name\_swap} vs.\ \texttt{abba}). Localised graph representations substantially exceed chance and edge/weight shuffling controls are near chance; prompt-only and raw patch-effect controls define strong comparison points. The current benchmark therefore establishes compression of slice-discriminative patching signal under explicit controls. Data compiled from our benchmark evaluation run.}
\label{tab:e3}
\end{table}

WL subtree features reach $0.912$, and localised edge-slot features reach $1.000$. A learned 2-layer GCN encoder (hidden dim $32$, trained end-to-end with the SVM target) reaches only $0.609$ in the linear-kernel setting and $0.583$ in the RBF setting, indicating that with $S=2$ slices and 32 bootstrap graphs per class the learned encoder underperforms hand-crafted localised features. Null controls (edge shuffle: $0.505$, weight shuffle: $0.519$) show that fixed-layout performance depends on preserving edge-slot assignments, not just marginal edge or weight statistics. The prompt-only baseline also reaches $1.000$, motivating the surface-balanced control below to separate graph signal from prompt artefacts.

For WL features, the linear kernel ($0.912$) outperforms RBF ($0.615$). We treat this as a benchmark observation: interpreting kernel spectrum or alignment requires a larger multi-slice dataset than the main $S=2$ comparison or the preliminary $S=4$ pilot.

In the historical \texttt{name\_swap} vs.\ \texttt{abba} benchmark, the gap between fixed-layout signed ($1.000$) and spectral ($0.500$) suggests that localised node/edge identity matters more than global graph shape. As shown next, this conclusion is benchmark-dependent: in the surface-balanced comparison, even compact global descriptors become perfectly classifiable.

The original \texttt{name\_swap} vs.\ \texttt{abba} comparison is confounded by prompt-level name counts. We therefore add \texttt{second\_subject\_swap}, whose corrupted prompts match \texttt{abba}'s target/distractor counts. In this control, prompt-only surface cues fall close to chance, while graph and raw patch-effect representations remain highly classifiable.

\begin{table}[!ht]
\centering
\scriptsize
\setlength{\tabcolsep}{3pt}
\begin{tabular*}{\textwidth}{@{\extracolsep{\fill}}llccccc@{}}
\toprule
\textbf{Benchmark} & \textbf{Dir.} & \textbf{Surface} & \textbf{Raw} & \textbf{WL} & \textbf{\shortstack{Fixed\\sign}} & \textbf{\shortstack{Edge\\shuffle}} \\
\midrule
\texttt{name}/\texttt{abba} & yes & $1.000\pm0.000$ & $1.000\pm0.000$ & $0.912\pm0.032$ & $1.000\pm0.000$ & $0.505\pm0.007$ \\
\texttt{abba}/\texttt{2nd-subj} & yes & $0.550\pm0.089$ & $1.000\pm0.000$ & $1.000\pm0.000$ & $1.000\pm0.000$ & $0.500\pm0.000$ \\
\texttt{abba}/\texttt{2nd-subj} & no & $0.550\pm0.089$ & $1.000\pm0.000$ & $1.000\pm0.000$ & $1.000\pm0.000$ & $0.500\pm0.000$ \\
\bottomrule
\end{tabular*}
\caption{Surface-balanced and direction-control results on GPT-2 IOI ($N=100$, $k=5$, CI graphs, seeds $\{7,42,123\}$). The surface-balanced benchmark removes the simple target/distractor count confound: the prompt-only baseline drops near chance, but graph and raw patch-effect classifiers remain perfect. Disabling the direction constraint does not change the main accuracies in this benchmark.}
\label{tab:surface_balanced}
\end{table}

Table~\ref{tab:cross_model} reports the same CI-graph protocol applied to DistilGPT-2 (6 layers, 768-dim, $|\node|=66$ residual nodes) under the historical \texttt{name\_swap} vs.\ \texttt{abba} pair. Localised edge-slot features (\emph{fixed sign}, \emph{hashed sign}, \emph{fixed binary}, \emph{coarse count}) and the raw patch-effect tensor remain near saturation in DistilGPT-2; WL subtree features drop from $0.912$ to $0.688$, indicating that subtree-level graph compression carries less signal in the smaller model. The GNN encoder baseline replicates at $0.609$ across both models. The combined picture is that localised edge-slot identity transfers across model sizes, while shape and subtree compressions are model-size dependent.
\begin{table}[!ht]
\centering
\scriptsize
\setlength{\tabcolsep}{4pt}
\begin{tabular*}{\textwidth}{@{\extracolsep{\fill}}llcc@{}}
\toprule
\textbf{Representation} & \textbf{Kernel} & \textbf{GPT-2 small} & \textbf{DistilGPT-2} \\
\midrule
Raw patch-effect tensor   & Linear & $1.000\pm0.000$ & $1.000\pm0.000$ \\
WL subtree                & Linear & $0.912\pm0.032$ & $0.688\pm0.130$ \\
Spectral                  & Linear & $0.500\pm0.000$ & $0.552\pm0.015$ \\
Graphlet                  & Linear & $0.760\pm0.124$ & $0.672\pm0.133$ \\
Fixed layout binary       & Linear & $0.990\pm0.015$ & $0.984\pm0.013$ \\
Fixed layout signed       & Linear & $1.000\pm0.000$ & $1.000\pm0.000$ \\
Hashed sign               & Linear & $0.995\pm0.007$ & $1.000\pm0.000$ \\
Coarse count              & Linear & $0.755\pm0.128$ & $0.911\pm0.053$ \\
GNN encoder + SVM         & Linear & $0.609\pm0.123$ & $0.609\pm0.109$ \\
Edge shuffle (fixed wgt.) & ---    & $0.505\pm0.007$ & $0.495\pm0.004$ \\
Weight shuffle (fixed wgt.) & ---  & $0.519\pm0.035$ & $0.523\pm0.008$ \\
\bottomrule
\end{tabular*}
\caption{Cross-model representation comparison on IOI \texttt{name\_swap} vs.\ \texttt{abba} (CI graphs, $N=100$, $k=5$, residual-stream nodes, seeds $\{7,42,123\}$). DistilGPT-2 has $|\node|=66$ vs.\ $|\node|=120$ for GPT-2 small. Localised edge-slot features and the raw patch-effect baseline transfer with negligible drop; WL subtree compression drops $\sim 22$ points. The GNN encoder is a 2-layer GCN with hidden dim $32$ trained end-to-end with the SVM target; it underperforms hand-crafted localised features in both models with the current $S=2$ slice budget.}
\label{tab:cross_model}
\end{table}

Table~\ref{tab:multislice_smoke} reports two $S=4$ multi-slice suites at $N=100$: an IOI/induction suite and a task-diverse suite that swaps \texttt{ioi:second\_subject\_swap} for \texttt{greater\_than:year\_swap}, spanning three task families. Localised edge-slot features reach $0.729$--$0.750$ in both suites versus chance $0.25$ and edge shuffle near $0.25$; raw patch effects sit at $0.758$ in both. The surface-cue baseline jumps from $0.500$ to $0.763$ when crossing task families, so multi-task accuracies must be read alongside surface controls.
\begin{table}[!ht]
\centering
\scriptsize
\setlength{\tabcolsep}{2pt}
\resizebox{\textwidth}{!}{%
\begin{tabular}{lccccccc}
\toprule
\textbf{Suite ($S=4$, $N=100$)} & \textbf{Surface} & \textbf{Raw} & \textbf{WL} & \textbf{\shortstack{Fixed\\sign}} & \textbf{\shortstack{Fixed\\weighted}} & \textbf{\shortstack{Edge\\shuffle}} & \textbf{\shortstack{Weight\\shuffle}} \\
\midrule
IOI/Ind & $0.500\pm0.037$ & $0.758\pm0.021$ & $0.708\pm0.015$ & $0.750\pm0.000$ & $0.750\pm0.000$ & $0.257\pm0.034$ & $0.549\pm0.123$ \\
IOI/GT/Ind & $0.763\pm0.031$ & $0.758\pm0.031$ & $0.594\pm0.077$ & $0.729\pm0.015$ & $0.750\pm0.000$ & $0.250\pm0.000$ & $0.472\pm0.013$ \\
\bottomrule
\end{tabular}%
}
\caption{Multi-slice evaluations on GPT-2 with CI graphs ($k=5$, 8 bootstrap graphs per slice, seeds $\{7,42,123\}$, chance $=0.25$). Top row (IOI/Ind): \texttt{ioi:\{abba,2nd-subj\}}, \texttt{induction\{,-late\}:token-swap}. Bottom row (IOI/GT/Ind) replaces \texttt{ioi:2nd-subj} with \texttt{greater\_than:year-swap}, covering three task families. Localised edge-slot features stay above edge-shuffle nulls in both suites; the higher surface-cue baseline in the diverse suite reflects vocabulary differences across task families.}
\label{tab:multislice_smoke}
\end{table}

Appendix Figure~\ref{fig:pareto_pca} shows the accuracy--dimensionality Pareto frontier (dominated by localised edge-slot features) and a PCA projection of WL embeddings; see Appendix~\ref{app:discussion} and~\ref{app:additional_experiments}.

\section{Limitations}
\label{sec:limitations}

The experiments are a controlled pilot, not a task-general benchmark. Binary IOI covers two slices, the $S=4$ IOI/induction and IOI/GT/induction suites stress-test three task families, and full DI graph construction is left as a scaling target ($O(|\node|^2)$ paired patches per prompt). Next: larger surface-balanced suites and full DI benchmarking.

\section{Conclusion}
\label{sec:conclusion}

We convert activation-patching interventions into patch-effect graphs classified by graph kernels, treating edge definition, representation, and similarity geometry as separable design axes. On GPT-2 small, localised CI-graph features achieve perfect IOI accuracy, including under a surface-balanced control where the prompt-only baseline collapses; $S=4$ IOI/induction and IOI/GT/induction suites plus a screened paired-patching study extend the protocol beyond binary IOI and validate CI/PC-ranked edges. Cross-model replication on DistilGPT-2 transfers localised edge-slot features with negligible drop, while WL subtree compression is model-size dependent. Strong raw and learned-encoder baselines impose a calibration principle: circuit-level claims must be evaluated against raw-tensor, surface-cue, and learned-encoder controls. Next: scaling full DI graphs and adopting surface-balanced, multi-slice, multi-task suites.

\bibliographystyle{plainnat}
\bibliography{references}

\appendix
\raggedbottom

\section{Additional Material}

\subsection{Additional Discussion}
\label{app:discussion}

\subsubsection{Summary of Evidence for Each Hypothesis}

\begin{table}[H]
\centering
\footnotesize
\begin{tabularx}{\textwidth}{@{}lL{3.0cm}L{2.3cm}X@{}}
\toprule
\textbf{Hyp.} & \textbf{Claim} & \textbf{Status} & \textbf{Evidence} \\
\midrule
H1 & Patch-effect graphs carry classifiable signal & \textbf{Supported in controlled setting} & Surface-balanced WL/fixed sign: $1.000$; four-slice $N=100$ WL/fixed sign: $0.708/0.750$ vs.\ surface cues $0.500$; raw patch effects remain strong \\
H2 & Kernel geometry shifts effective similarity space & \textbf{Requires larger $S$} & Current $S=2$ and $S=4$ runs establish protocol viability; spectrum/alignment analysis needs broader slice coverage \\
H3 & Edge construction changes graph representations & \textbf{Screened-DI validated} & CI and PC differ under WL; PC remains strong with fixed edge-slot signs; screened DI validates selected candidates \\
\bottomrule
\end{tabularx}
\caption{Hypothesis status after IOI pilot experiments and the preliminary four-slice evaluation. Current evidence supports controlled compression of classifiable patching signal, including under a surface-balanced corruption pair; broader multi-family suites are the next step for task-general circuit fingerprinting.}
\label{tab:hypotheses}
\end{table}

\subsubsection{What the Pilot Results Mean for Graph Construction}

CI graphs are strongly classifiable in both IOI binary benchmarks: WL+linear achieves $0.912$ on \texttt{name\_swap} vs.\ \texttt{abba} and $1.000$ on the surface-balanced comparison. PC behaves differently: fixed edge-slot signs remain highly predictive, while WL+linear drops to $0.552$ on the historical benchmark and recovers to $0.917$ on the surface-balanced benchmark. This indicates that the edge-construction rule materially changes graph topology as seen by WL features, with the preferred construction depending on representation and benchmark. The screened DI validation shows that CI/PC-ranked candidates have stronger destination activation influence than random or low-rank candidates under paired patching, giving targeted evidence that the ranked edges correspond to causally active locations.

\subsubsection{What the Pilot Results Mean for Kernel Choice}

At $S=2$ slices, the linear kernel outperforms RBF for WL features. This is a useful benchmark observation, while full kernel-spectrum or alignment analysis requires more underlying slices. The preliminary four-slice evaluation shows that the classifier protocol can run beyond binary IOI and that graph features remain classifiable, motivating a larger multi-slice dataset for kernel-geometry analysis.

\subsubsection{The Surprising Efficacy of Raw Patch-Effect Representations}

A key diagnostic finding is that raw patch-effect vectors are highly competitive with graph baselines. The surface-balanced IOI control reduces the simple prompt-count confound, and the preliminary four-slice evaluation shows graph signal beyond a single binary IOI comparison. This positions graph construction as structured compression and edge-level auditability over a strong raw-tensor baseline.

\subsubsection{Scalability}

The primary scalability bottleneck is paired patching for DI graphs: $O(|\node|^2)$ forward passes per prompt. For GPT-2 with $|\node|=120$ residual nodes, this is $\binom{120}{2} = 7{,}140$ ordered pairs per prompt, feasible for $N=100$ prompts but requiring careful activation caching. For larger models, attribution patching \cite{nanda2023attribution} can serve as a cheap screening step to identify candidate pairs before expensive paired patching.

Graph embeddings (WL, spectral, graphlet) are fixed-dimension and do not grow quadratically with model size. Hashed fixed-layout features use a fixed $d=1{,}024$ coordinates regardless of $|\node|$, making them particularly suitable for larger-model extensions.

\begin{figure}[H]
\centering
\includegraphics[width=0.75\textwidth]{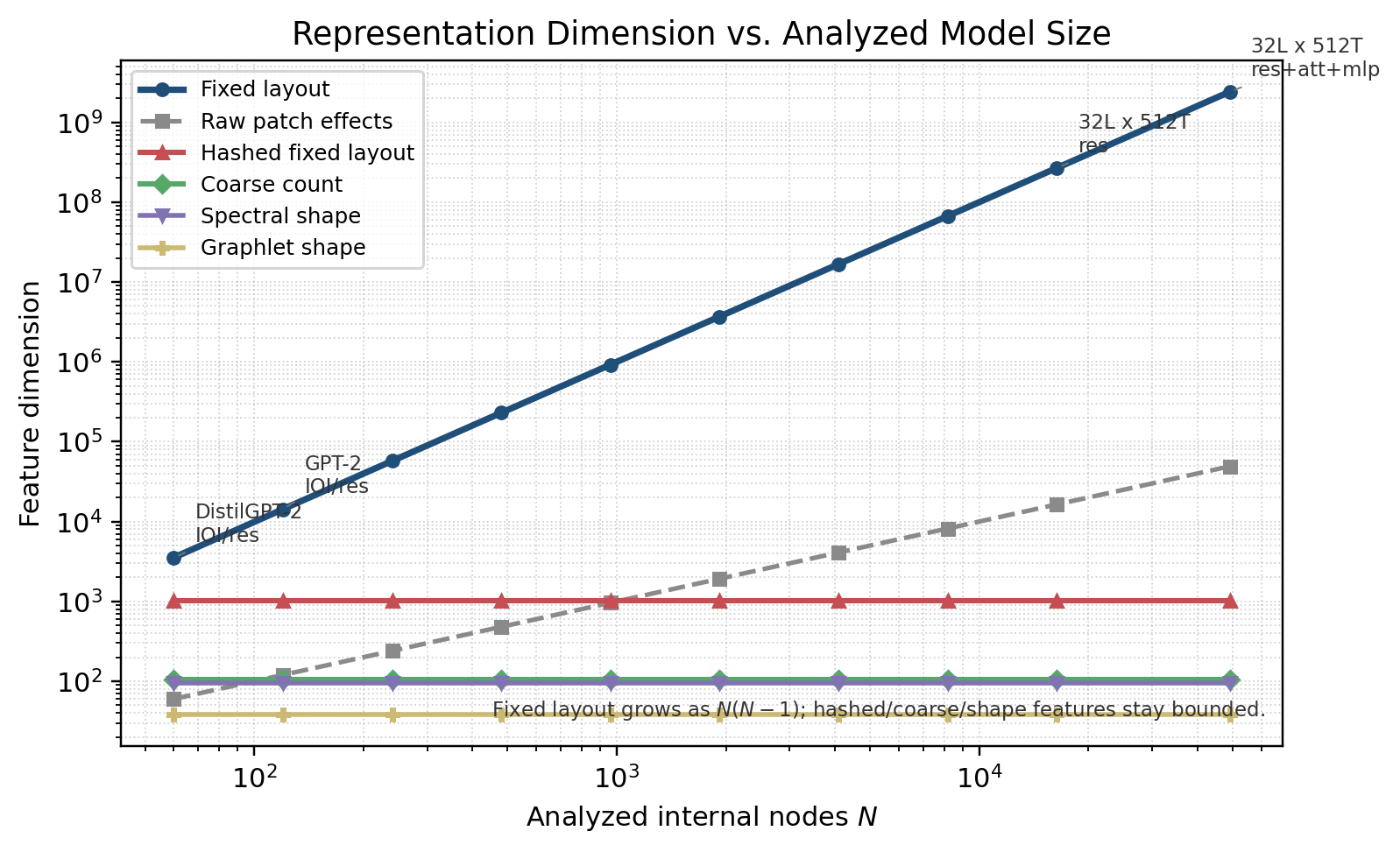}
\caption{Feature dimensionality as the analyzed node set grows ($N = L \times T \times C$). Fixed-layout features grow as $N(N-1)$ (quadratic); raw patch-effect vectors grow linearly. Hashed fixed-layout, coarse count, spectral, and graphlet features remain bounded regardless of model size, making them the scalable alternatives for larger transformers.}
\label{fig:scaling}
\end{figure}

\subsection{Additional Experimental Details}
\label{app:additional_experiments}

\subsubsection{Embedding Visualisations}

\begin{figure}[!t]
\centering
\begin{subfigure}[b]{0.98\textwidth}
  \includegraphics[width=\textwidth]{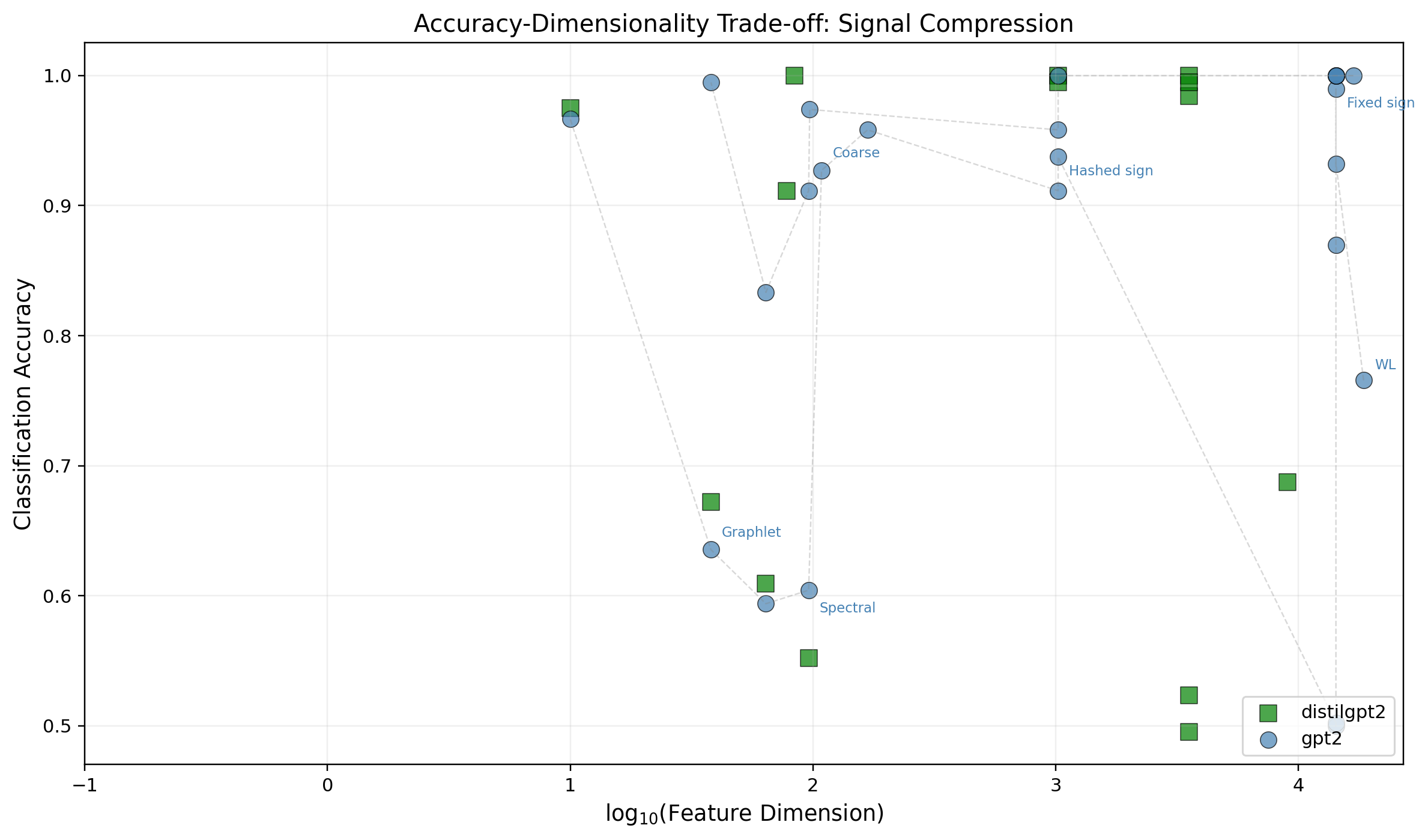}
  \caption{Accuracy--dimensionality trade-off (Pareto frontier)}
  \label{fig:pareto}
\end{subfigure}
\par\medskip
\begin{subfigure}[b]{0.76\textwidth}
  \includegraphics[width=\textwidth]{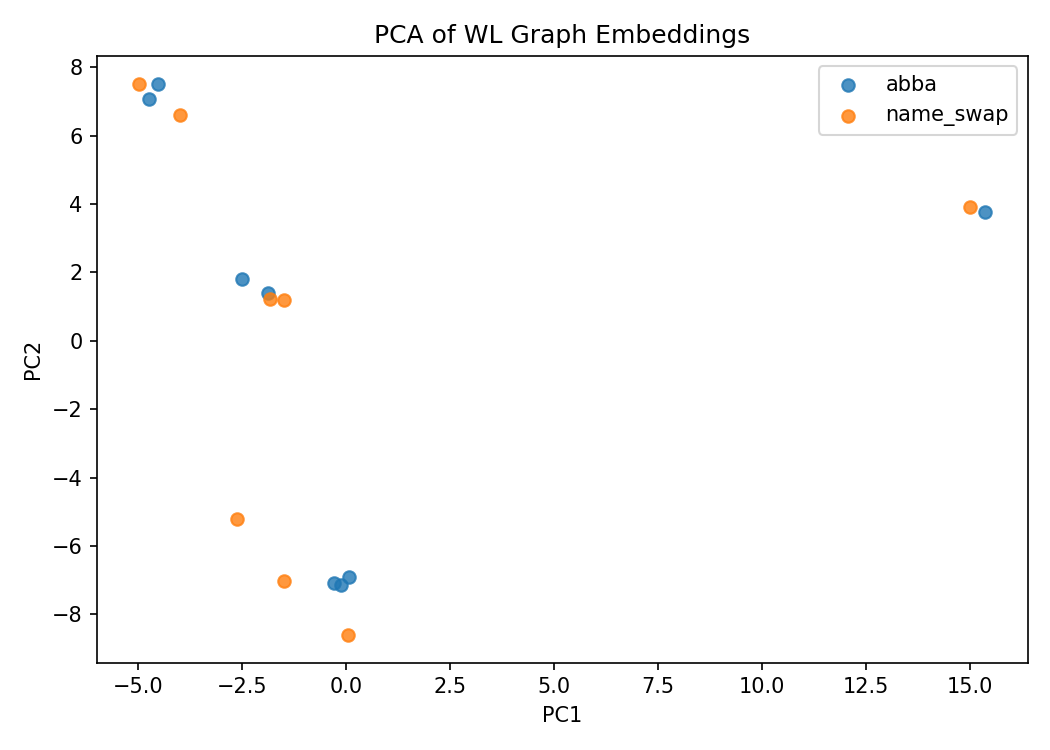}
  \caption{PCA of WL graph embeddings}
  \label{fig:pca}
\end{subfigure}
\caption{Left: accuracy vs.\ $\log_{10}$ feature dimension for graph representations on GPT-2 (circles) and DistilGPT-2 (squares) from the compression sweep. The Pareto frontier is dominated by localised edge-slot features; global shape descriptors lie well below it. Right: 2D PCA projection of WL subtree embeddings for the $2\times 32$ bootstrap graphs (32 per slice); colours indicate corruption type (\texttt{abba} vs.\ \texttt{name\_swap}). This visualisation is intended as qualitative support for the classifier results.}
\label{fig:pareto_pca}
\end{figure}

Figure~\ref{fig:pareto_pca} reports the compression-sweep visualisations used to contextualise the appendix results.

\subsubsection{Prompt Budget}
\label{sec:prompt_budget}

Table~\ref{tab:n500} shows results at $N=500$ examples per slice from a single-seed scalability run. Localised representations saturate ($1.000$), and spectral features improve substantially. Because a prompt-only surface-cue classifier reaches $1.000$ in the current $N=100$ historical control, the $N=500$ run is best read as a prompt-budget and compression check rather than a mechanistic-depth claim.

\begin{table}[H]
\centering
\small
\begin{tabular}{lcc}
\toprule
\textbf{Representation} & \textbf{$N=100$} & \textbf{$N=500$} \\
\midrule
WL linear           & $0.9115$ & $1.0000$ \\
Fixed layout signed & $1.0000$ & $1.0000$ \\
Hashed sign         & $0.9948$ & $1.0000$ \\
Coarse count        & $0.7552$ & $0.9740$ \\
Spectral linear     & $0.5000$ & $0.9115$ \\
Graphlet linear     & $0.7604$ & $0.9948$ \\
\bottomrule
\end{tabular}
\caption{Effect of prompt budget ($N$ examples per slice) on classification accuracy (IOI, GPT-2, CI graphs). The $N=100$ column reports the current three-seed CI run; the $N=500$ column is a single-seed scalability run and should be read as descriptive. Localised representations are near-saturated at $N=100$; shape descriptors improve at $N=500$, consistent with additional prompt budget improving graph-compression stability.}
\label{tab:n500}
\end{table}

\subsubsection{E4: Robustness Ablations}
\label{sec:e4}

Table~\ref{tab:n500} characterises accuracy as a function of $N \in \{100, 500\}$ examples per slice. Localised representations are already near-saturated at $N=100$; shape descriptors are the most sensitive to sample size.

The main protocol uses 32 bootstrap graphs per slice with 75\% resampling. To check that the surface-balanced result is not an artefact of that exact choice, we reran the CI surface-balanced comparison with only 8 bootstrap graphs per slice and resampling fractions of 50\% and 100\%. Table~\ref{tab:bootstrap_smoke} shows that WL and fixed-sign accuracy remain saturated, while edge shuffling remains at chance. This is a useful sensitivity check, not a replacement for a full bootstrap hyperparameter sweep.

\begin{table}[H]
\centering
\small
\begin{tabular}{ccccc}
\toprule
\textbf{Graphs/slice} & \textbf{Sample frac.} & \textbf{WL} & \textbf{Fixed sign} & \textbf{Edge shuffle} \\
\midrule
32 & $0.75$ & $1.0000 \pm 0.0000$ & $1.0000 \pm 0.0000$ & $0.5000 \pm 0.0000$ \\
8  & $0.50$ & $1.0000 \pm 0.0000$ & $1.0000 \pm 0.0000$ & $0.5000 \pm 0.0000$ \\
8  & $1.00$ & $1.0000 \pm 0.0000$ & $1.0000 \pm 0.0000$ & $0.5000 \pm 0.0000$ \\
\bottomrule
\end{tabular}
\caption{Fast bootstrap sensitivity check on the surface-balanced GPT-2 IOI comparison (\texttt{abba} vs.\ \texttt{second\_subject\_swap}, $N=100$, $k=5$, CI graphs, seeds $\{7,42,123\}$). Main accuracies do not depend on the exact 32-graph/75\% bootstrap choice in this saturated benchmark.}
\label{tab:bootstrap_smoke}
\end{table}

The main results use the direction constraint ($u \prec v$ by layer then token). In the surface-balanced IOI control, disabling the direction constraint leaves WL+linear, fixed sign, raw patch-effect, spectral, and graphlet accuracies at $1.0000$ (Table~\ref{tab:surface_balanced}). Thus the current surface-balanced classification result is not dependent on the direction constraint, although direction may matter in less saturated multi-slice benchmarks.

Edge shuffle and weight shuffle both reduce fixed-layout weighted accuracy to near chance (Table~\ref{tab:e3}), confirming the signal depends on preserving edge-slot identity rather than marginal edge or weight statistics.

Prompt-only surface cues reach $1.0000 \pm 0.0000$ on \texttt{name\_swap} vs.\ \texttt{abba}, but only $0.5500 \pm 0.0890$ on the surface-balanced \texttt{abba} vs.\ \texttt{second\_subject\_swap} comparison. Raw patch-effect vectors remain at $1.0000 \pm 0.0000$ in both settings. This separates two issues: the original benchmark has a surface-count shortcut, and graph construction currently matches strong raw-tensor baselines while adding structured compression and edge-level auditability.

\subsubsection{E5: Graph Construction Ablation}
\label{sec:e5}

Test whether the edge-construction rule changes downstream classifiability. The current fast ablation compares CI against PC, while screened DI provides targeted causal validation.

PC is now implemented and evaluated on both the historical and surface-balanced IOI comparisons. Exhaustive DI graph classification is treated as a scaling target because it requires paired-node patching at full graph scale.

\begin{table}[H]
\centering
\scriptsize
\setlength{\tabcolsep}{3pt}
\begin{tabular*}{\textwidth}{@{\extracolsep{\fill}}llcccc@{}}
\toprule
\textbf{Benchmark} & \textbf{Construction} & \textbf{\shortstack{WL+linear\\Acc.}} & \textbf{\shortstack{Fixed sign\\Acc.}} & \textbf{\shortstack{Surface\\cue}} & \textbf{Cost} \\
\midrule
\texttt{name}/\texttt{abba} & Co-influence (CI) & $0.912\pm0.032$ & $1.000\pm0.000$ & $1.000\pm0.000$ & $O(|\node|)$ fwd. \\
\texttt{name}/\texttt{abba} & Partial corr.\ (PC) & $0.552\pm0.058$ & $0.990\pm0.015$ & $1.000\pm0.000$ & $O(|\node|^3)$ off. \\
\texttt{abba}/2nd-subj & Co-influence (CI) & $1.000\pm0.000$ & $1.000\pm0.000$ & $0.550\pm0.089$ & $O(|\node|)$ fwd. \\
\texttt{abba}/2nd-subj & Partial corr.\ (PC) & $0.917\pm0.118$ & $1.000\pm0.000$ & $0.550\pm0.089$ & $O(|\node|^3)$ off. \\
Any & Direct influence (DI) & \textit{screened only} & \textit{screened only} & --- & $O(|\node|^2)$ fwd. \\
\bottomrule
\end{tabular*}
\caption{Graph construction ablation on IOI with GPT-2 ($N=100$, $k=5$, seeds $\{7,42,123\}$). PC preserves strong fixed edge-slot performance in both benchmarks. WL+linear is near chance for PC on the historical surface-confounded benchmark, but improves on the surface-balanced benchmark; therefore the robust claim is construction sensitivity, not a universal CI$>$PC ordering. DI is evaluated through the screened paired-patching validation in Table~\ref{tab:screened_di}.}
\label{tab:e5}
\end{table}

\begin{table}[H]
\centering
\scriptsize
\setlength{\tabcolsep}{4pt}
\begin{tabular*}{\textwidth}{@{\extracolsep{\fill}}lccccc@{}}
\toprule
\textbf{Candidate screen} & \textbf{Edges} & \textbf{$I$ mean$\pm$std} & \textbf{$I$ FDR} & \textbf{$M$ mean$\pm$std} & \textbf{Nec.\ mean$\pm$std} \\
\midrule
Top CI & 50 & $0.373\pm0.059$ & 50/50 & $0.018\pm0.027$ & $0.068\pm0.146$ \\
Top PC & 50 & $0.305\pm0.187$ & 45/50 & $0.053\pm0.178$ & $-0.213\pm0.375$ \\
Random & 50 & $0.122\pm0.253$ & 29/50 & $0.073\pm0.716$ & $0.052\pm0.303$ \\
Low-rank CI & 50 & $0.102\pm0.153$ & 33/50 & $0.112\pm0.454$ & $0.000\pm0.000$ \\
\bottomrule
\end{tabular*}
\caption{Screened direct-influence candidate validation on GPT-2 small residual-stream nodes. Candidate edges are selected on a discovery split from CI, PC, random, or low-rank CI screens and evaluated on held-out prompts with paired patching. $I$ is activation influence at the destination node, $M=R(\{u,v\})-R(\{v\})$ is output-level mediation, and FDR counts use BH correction at $\alpha=0.05$.}
\label{tab:screened_di}
\end{table}

\subsubsection{E6: Kernel Geometry Analysis}
\label{sec:e6}

Eigenvalue spectrum, kernel alignment (CKA), and kernel-PCA projections require a meaningful kernel matrix $K \in \R^{S \times S}$ over multiple underlying slices. The main IOI benchmarks have $S=2$ slices, and the new $S=4$ IOI/induction preliminary multi-slice evaluation is useful for checking classifiability. We therefore treat H2 as a larger-suite analysis target rather than over-interpreting the present small kernel matrices.

A valid kernel-geometry analysis needs additional task/corruption families, surface-balanced slice definitions, and enough slices to make the spectrum and alignment of $K$ interpretable.

\subsubsection{Metric and Control Glossary}
\label{sec:metric_control_glossary}

Table~\ref{tab:metric_glossary} summarises the quantities used across the experiments. Its purpose is only to make the evaluation protocol auditable; formal definitions for the graph constructions are in Section~\ref{sec:graph}.

\begin{table}[!htbp]
\centering
\scriptsize
\setlength{\tabcolsep}{3pt}
\begin{tabularx}{\textwidth}{@{}lL{2.35cm}X@{}}
\toprule
\textbf{Quantity} & \textbf{Role} & \textbf{Interpretation} \\
\midrule
$E_u$ & Node patch effect & Logit restoration from patching node $u$ in the corrupted run; the raw signal before graph construction. \\
CI / PC / DI & Edge construction & CI uses patch-effect correlation; PC removes linear shared-cause structure; DI is the paired-patching causal target. \\
WL / fixed / spectral / graphlet & Graph representation & Alternative compressions of each slice graph, ranging from local edge-slot identity to global graph-shape descriptors. \\
Prompt-only / raw tensor & Non-graph controls & Tests whether classification can be solved from surface text or from unstructured patch effects without graph construction. \\
Edge / weight shuffle & Null controls & Tests whether fixed-layout performance depends on preserving edge identity rather than marginal edge or weight statistics. \\
Bootstrap accuracy & Reported outcome & Accuracy on example-disjoint bootstrap graphs; it measures slice-level generalisation within the reported protocol. \\
\bottomrule
\end{tabularx}
\caption{Compact glossary of the metrics, representations, and controls used in the experimental tables.}
\label{tab:metric_glossary}
\end{table}

\subsubsection{Prompt Data and Splits}
\label{sec:prompt_data_splits}

All reported data are synthetic prompts generated from fixed templates and seeds. The main IOI runs use $N=100$ paired examples per slice and seeds $\{7,42,123\}$; the prompt-budget check adds a single-seed $N=500$ run. Training and test graphs are built from disjoint example pools, and bootstrap graphs resample within the relevant pool only. The surface-balanced comparison, \texttt{abba} vs.\ \texttt{second\_subject\_swap}, is included because it matches the target/distractor name-count structure that makes the historical \texttt{name\_swap} vs.\ \texttt{abba} benchmark surface-separable.

\subsubsection{Threats to Causal Interpretation}
\label{sec:threats_causal_interpretation}

The experiments are designed to separate representation quality from stronger claims about causal discovery. The binary IOI setting has only two underlying slices, so we report bootstrap accuracy as slice-discriminative evidence and add the four-slice pilot to check transfer beyond IOI. Raw patch-effect vectors remain a strong baseline; this is informative because the graph pipeline is intended to compress and organise patching signal while making the value of additional structure auditable. Finally, screened DI validates that CI/PC-selected locations are causally active under paired patching, while a full DI graph benchmark remains the natural next step for testing causal-mediation superiority directly.

\subsection{Reproducibility and Responsible Release}
\label{app:reproducibility_release}

The experiments use synthetic prompt templates and existing GPT-2 weights; no private data source is required and no fixed data bundle is released. Prompt pairs and patch-effect artifacts are generated on demand for each run from the provided templates, seeds, and scripts. A public repository and supplementary archive are planned for release with the implementation, experiment scripts, generated figure artifacts, and prompt-generation code. The planned repository license for the released code and generated artifacts is MIT.

The work is foundational mechanistic interpretability research. Potential positive impact comes from improving tools for auditing internal model computations and comparing circuit-level behaviour across controlled tasks. Potential negative impact is indirect: better interpretability tools can also help capable actors diagnose and adapt model behaviours, including behaviours that might be undesirable in deployed systems. The paper does not release new model weights or scraped data, and the experiments use synthetic prompts and existing GPT-2 weights, limiting direct deployment risk.

GPT-2 small is credited via \citet{radford2019language}; the IOI task and circuit reference are credited via \citet{wang2023localizing}. The implementation relies on standard open-source Python scientific and ML packages. The released project code and generated artifacts are planned under the MIT License. The paper does not redistribute GPT-2 weights in the manuscript source; users should obtain them through their standard provider and comply with the corresponding provider terms and model license. Synthetic prompts generated by this work do not contain personal or private data.

\end{document}